\documentclass[sigconf]{acmart}

\AtBeginDocument{%
  }
    
\setcopyright{acmlicensed}
\copyrightyear{2025}
\acmYear{2025}
\acmDOI{XXXXXXX.XXXXXXX}

\acmSubmissionID{287}

\usepackage{multirow,booktabs}
\usepackage{makecell}
\usepackage{algorithm}
\usepackage{algpseudocode}

\usepackage{bm} 
\usepackage{amsmath}
\usepackage{hyperref}
\newcommand{\mat}[1]{\mathbf{#1}}

\newcommand{\fthref}[1]{\footnote{\href{{#1}}{{#1}}}}

\begin{document}
\title{Efficient Implicit Neural Compression of Point Clouds via Learnable Activation in Latent Space}

\author{Yichi Zhang}
\affiliation{%
  \institution{Zhejiang University}
  \city{Hangzhou}
  \country{China}
}
\email{yichizhang@zju.edu.cn}

\author{Qianqian Yang}
\affiliation{%
  \institution{Zhejiang University}
  \city{Hangzhou}
  \country{China}}
\email{qianqianyang20@zju.edu.cn}

\renewcommand{\shortauthors}{Zhang et al.}


\begin{abstract}

Implicit Neural Representations (INRs), also known as neural fields, have emerged as a powerful paradigm in deep learning, parameterizing continuous spatial fields using coordinate-based neural networks. In this paper, we propose \textbf{PICO}, an INR-based framework for static point cloud compression. Unlike prevailing encoder-decoder paradigms, we decompose the point cloud compression task into two separate stages: geometry compression and attribute compression, each with distinct INR optimization objectives. Inspired by Kolmogorov-Arnold Networks (KANs), we introduce a novel network architecture, \textbf{LeAFNet}, which leverages learnable activation functions in the latent space to better approximate the target signal's implicit function. By reformulating point cloud compression as neural parameter compression, we further improve compression efficiency through quantization and entropy coding. Experimental results demonstrate that \textbf{LeAFNet} outperforms conventional MLPs in INR-based point cloud compression. Furthermore, \textbf{PICO} achieves superior geometry compression performance compared to the current MPEG point cloud compression standard, yielding an average improvement of $4.92$ dB in D1 PSNR. In joint geometry and attribute compression, our approach exhibits highly competitive results, with an average PCQM gain of $2.7 \times 10^{-3}$.

\end{abstract}

\ccsdesc[300]{Computing methodologies~Shape representations}
\ccsdesc[300]{Computing methodologies~Appearance and texture representations}
\ccsdesc[300]{Computing methodologies~Image compression}

\keywords{Point Cloud Compression, Implicit Neural Representation, Learnable Activation Function}


\maketitle

\section{Introduction}


Point clouds have become a widely adopted data format for representing 3D objects and environments, playing a crucial role in applications such as autonomous driving~\cite{li2020deep, cui2021deep}, augmented and virtual reality (AR/VR)~\cite{wang2023pointshopar, lim2022point}, and embodied intelligence~\cite{qi2024shapellm}. With the advancement and widespread adoption of LiDAR sensing technology~\cite{raj2020survey}, capturing large-scale point clouds with high-resolution spatial and attribute information has become increasingly feasible. 
However, raw point cloud data is characterized by massive scale, spatial sparsity, and an inherent lack of structural coherence, posing significant challenges for storage and transmission due to high memory and bandwidth requirements~\cite{graziosi2020overview}.  These challenges highlight the urgent need for efficient point cloud compression (PCC) solutions to enable practical deployment in real-world scenarios.

The Moving Picture Experts Group (MPEG) spearheaded PCC standardization in 2017~\cite{schwarz2018emerging}, culminating in 2020 with two framework specifications: Geometry-based PCC (G-PCC) and Video-based PCC (V-PCC)~\cite{graziosi2020overview}. G-PCC processes native 3D data through adaptive voxelization with level-of-detail segmentation that enables progressive geometric compression. In contrast, V-PCC performs geometry-aware planar projections to transform 3D data into 2D atlas representations compatible with mature video codecs.

Building on advances in learned image compression, recent work has extended data-driven approaches to PCC~\cite{huang20193d, quach2020improved, que2021voxelcontext, zhang2024deeppcc}. Current learning-based PCC frameworks typically adopt autoencoder-style architectures where an encoder network projects input points to a latent code, while a decoder reconstructs the geometry and attributes. Although these methods achieve rate-distortion advantages over conventional codecs, they face two fundamental limitations: (1) parametrization critically dependent on training data statistics, and (2) limited cross-distribution generalization due to the parametric bottleneck~\cite{quach2022survey}.


Implicit neural representations (INRs), colloquially termed neural fields, have emerged as a research frontier in deep learning that provides a coordinate-driven framework for continuous signal parameterization~\cite{sitzmann2020implicit, ramasinghe2022beyond, saragadam2023wire}. These methods fundamentally redefine field representation through neural networks that map spatial coordinates to signal attributes. 
The paradigm shift began with Neural Radiance Fields (NeRF)~\cite{mildenhall2021nerf}, which established a seminal framework for 3D scene reconstruction and novel view synthesis through differentiable volume rendering. Subsequent extensions demonstrate INR's versatility: 2D image representations achieve memory-efficient compression via coordinate-based weight sharing~\cite{strumpler2022implicit}, while video INRs enable temporal consistency through latent space dynamics~\cite{chen2021nerv}. This reconstruction-as-optimization perspective has further inspired multimodal compression techniques~\cite{wang2022clip, ballerini2024connecting, ruan2024implicit} that jointly optimize network parameters across data modalities.


In this paper, we propose a systematic INR framework for joint geometry and attribute compression of LiDAR-acquired point clouds, \textbf{PICO}~(\textsc{\underline{P}oint cloud \underline{I}mplicit neural \underline{CO}mpression}). Our methodology rethinks PCC as a two-stage task decomposition: geometry compression followed by attribute compression. Each stage is formulated with domain-specific objectives to maximize coding efficiency.  
The geometry compression stage leverages a neural network to predict voxel occupancy probabilities in the voxelized space, utilizing dynamic thresholds to achieve high-quality geometry reconstruction. For attribute compression, another neural network restores surface properties including spatially-varying normals, color information, and material characteristics directly on the reconstructed geometry.

Inspired by Kolmogorov-Arnold Networks (KANs)~\cite{liu2025kan}, we introduce \textbf{LeAFNet} (\textsc{\underline{Le}arnable} \textsc{\underline{A}ctivation} \textsc{\underline{F}unction} \textsc{\underline{Net}work}), a novel architecture for INR-based PCC. Our framework incorporates parametric activation operators in latent representation learning to substantially improve implicit field approximation accuracy. The learnable activation function adaptively capture multi-scale features critical for sparse point cloud representations. LeAFNet achieves competitive reconstruction quality while requiring much fewer parameters than standard multilayer perceptrons, demonstrating superior efficiency for compression-oriented implementations.

Our framework recasts the standardized PCC pipeline into a neural parameter optimization framework. Upon INR convergence, we execute parameter quantization followed by entropy coding, yielding a compact binary bitstream for storage or transmission. The computationally efficient decompression phase reconstructs the complete point cloud through coordinate queries alone, preserving structural integrity across arbitrary sampling resolutions.


PICO achieves efficient compression through instance-specific optimization on individual point clouds, leveraging neural networks' approximation power while eliminating redundancy via decoupled geometry and attribute compression. The lightweight architecture of LeAFNet enables full parameter utilization without compromising model efficiency.
PICO implements adaptive L1 regularization for precise rate control, circumventing the training instability caused by entropy estimation in autoencoder methods. Notably, PICO supports spatial block-wise compression for large-scale scenes, enabling parallel compression that outperforms traditional approaches in reconstruction scalability.
Furthermore, PICO's compatibility with advancing quantization and model encoding techniques ensures continuous performance improvements, establishing practical value for real-world deployment.

We benchmark PICO against the MPEG standardization anchors G-PCC and V-PCC on \texttt{8iVFB}. For geometry compression, our method achieves 53.54\% BD-BR savings with 4.92 dB BD-PSNR gains. In joint geometry-attribute compression, PICO reduces BD-BR by 42.71\% and improve BD-PCQM by $2.70\times10^{-3}$. These results demonstrate the superiority of PICO. 

Our key contributions are as follows:

(1) We propose \textbf{PICO}, a novel PCC framework that decouples compression into geometry and attribute components, achieving enhanced computational efficiency and superior scalability.

(2) We present \textbf{LeAFNet}, a novel INR-based backbone with learnable activation function that enhances implicit function approximation ability while preserving parameter-efficient design.

(3) We re-express PCC as neural network compression task, integrating advanced techniques including precision quantization and entropy-constrained encoding into the compression pipeline.

(4) We validate PICO through extensive evaluations, demonstrating improved compression performance against MPEG standards.

\section{Related Work}

\subsection{Point Cloud Compression}


The MPEG 3D Graphics Coding Group has established the PCC standard, introducing two methods, G-PCC and V-PCC~\cite{graziosi2020overview}, to meet the increasing demand for efficient data representation and storage. 
G-PCC directly encodes both geometric and attribute information in 3D space. For geometry compression, G-PCC uses an octree structure to represent voxelized point cloud geometry, where each node is assigned a binary label, forming a binary string that is compressed using entropy coding. It also provides a Trisoup encoding technique for geometry, which approximates the object's surface using triangle meshes and performs especially well at low bitrates. For attribute compression, G-PCC employs geometry-based linear transformations, including the region-adaptive hierarchical transform (RAHT)~\cite{de2016compression}, which predicts higher-level attribute values of the octree based on lower-level values. 
On the other hand, V-PCC decomposes 3D point clouds through planar parameterization, generating atlas-organized geometry patches and attribute maps. These components are systematically packed into 2D video frames and compressed via HEVC~\cite{sullivan2012overview} with geometry-color separation, where motion-compensated prediction optimizes inter-patch correlation while preserving structural continuity during reconstruction.

Recent advances in deep learning have significantly advanced PCC through autoencoder-based frameworks. These methods encode point clouds into latent representations and reconstruct them via decoding, where learned entropy models constrain the latent distribution to optimize rate-distortion trade-offs. Addressing the intrinsic irregularity of point cloud data, researchers have proposed distinct architectural paradigms. 
Voxel-based methods~\cite{wang2021lossy, wang2021multiscale, wang2022sparse} implement hybrid 3D convolutions (dense/sparse) for geometry compression on discretized volumetric representations. Point-wise compression frameworks~\cite{huang20193d, sheng2021deep} inspired by PointNet~\cite{qi2017pointnet}  architectures process raw point coordinates directly, avoiding voxelization artifacts.
While surpassing traditional image codecs in rate-distortion performance, current learning-based approaches still struggle with cross-domain generalization and scalability limitations~\cite{quach2022survey}, particularly when handling large-scale scenes with variable density distributions. 

\subsection{Implicit Neural Representation}

Implicit neural representation (INR)~\cite{ramasinghe2022beyond, saragadam2023wire, sitzmann2020implicit} parameterizes continuous multidimensional signals through coordinate-based neural networks. 
Given input features be $\bm{x} \in \mathbb{R}^{d_i}$ and output features be $\bm{y} \in \mathbb{R}^{d_o}$. Then, the INR can be expressed as $f(\cdot;\Theta): \mathbb{R}^{d_i} \to \mathbb{R}^{d_o}$, where $\Theta$ represents the neural network parameters. These parameters can be determined by minimizing the error $\mathcal{L}(\cdot, \cdot)$ between the predicted output and the ground truth, which can be formulated as an optimization problem:
\begin{equation}
    \underset{\Theta}{\text{argmin}} \ \mathbb{E}_{\bm{x}, \bm{y} \sim P_{\text{data}}} \big[ \mathcal{L}(f(\bm{x};\Theta), \bm{y}) \big].
\end{equation}  

Numerous studies have explored compressing multimodal data using INR. Methods such as COIN~\cite{dupont2021coin, strumpler2022implicit, dupont2022coin++} propose using MLPs and meta-learning to map image pixel coordinates to corresponding RGB values, while employing meta-learning techniques to accelerate the compression process on large-scale datasets. Similarly, approaches like IPF~\cite{zhang2021implicit} utilize inter-frame information and INRs for video compression. All these methods provide valuable insights and experiences for using INRs in PCC.

\subsection{Learnable Activation Functions}

While specialized activation functions mitigate spectral bias in INRs~\cite{heidari2024single}, they are highly sensitive to initialization strategies and hyperparameter selection, introducing new challenges during training~\cite{liu2024finer}. Therefore, it is crucial to develop adaptive functions that can effectively handle nonlinearities and complex frequency distributions. Recent research has shifted towards using a portion of the network parameters for learnable activation functions. These learnable activation functions are not fixed but optimized during training to dynamically adapt to the data and learning process~\cite{goyal2019learning, bingham2022discovering}, providing a more powerful INR to fit implicit functions.

Kolmogorov-Arnold Network~\cite{liu2025kan} is grounded in the Kolmogorov-Arnold representation theorem, which states that any multivariate continuous function $f$ defined on a bounded domain can be expressed as a finite composition of continuous functions of a single variable combined with binary addition operations. 
More specifically, the theorem asserts that for a continuous function $f: [0,1]^n \rightarrow \mathbb{R}$, there exist univariate continuous functions $\Phi_{q}$ and $\phi_{q,p}$ such that:
\begin{align}
f(x_1, x_2, \dots, x_n) = \sum_{q=1}^{2n+1} \Phi_{q}\left( \sum_{p=1}^{n} \phi_{q,p}(x_p) \right)
\end{align}
where $\Phi_q:\mathbb{R}\rightarrow\mathbb{R}$, $\phi_{q,p}:[0,1]\rightarrow\mathbb{R}$ are trainable functions. 
KAN~\cite{liu2025kan} proposed to combine B-spline and SiLU~\cite{jocher2021ultralytics} to implement these trainable functions. 
\begin{align}
&\phi(x) = w_b\texttt{silu}(x) + w_s\texttt{spline}(x).
\end{align}
By abstracting away the specific forms of these functions, KAN can be formalized into the following matrix representation:
\begin{align}
{\rm KAN}(\mat{x}) = (\mat{\Phi}_{L-1}\circ \mat{\Phi}_{L-2}\circ\cdots\circ\mat{\Phi}_{1}\circ\mat{\Phi}_{0})\mat{x}. 
\end{align}
Here, the forward process of KAN can be expressed as
\begin{align}
    \mat{x}_{l+1} = \left[\begin{matrix}
        \phi_{l,1,1}(\cdot) & \phi_{l,1,2}(\cdot) & \cdots & \phi_{l,1,n_{l}}(\cdot) \\
        \phi_{l,2,1}(\cdot) & \phi_{l,2,2}(\cdot) & \cdots & \phi_{l,2,n_{l}}(\cdot) \\
        \vdots & \vdots & & \vdots \\
        \phi_{l,n_{l+1},1}(\cdot) & \phi_{l,n_{l+1},2}(\cdot) & \cdots & \phi_{l,n_{l+1},n_{l}}(\cdot) \\
    \end{matrix}\right]
    \mat{x}_{l} = \mat{\Phi}_l \mat{x}_{l}. 
\end{align}

Although KAN exhibits outstanding performance in fitting scientific functions and capturing high-frequency signal variations, its application to complex deep learning tasks remains challenging due to the non-smoothness of the inner function and its low computational efficiency~\cite{dahal2025efficiency}. Recently, researchers have explored replacing the B-spline function with alternative formulations and improving the KAN architecture to achieve more accurate solutions with lower time complexity. For instance, Fourier KAN~\cite{xu2024fourierkan} replaces $\phi(x)$ with Fourier series, while fKAN~\cite{aghaei2025fkan} adopts Jacobi basis functions.

\section{Methodology}

\subsection{Preliminaries}\label{sec:preliminaries}

In a hypothetical space voxelized with $N$-bit resolution, the entire voxelized space comprises $2^N \times 2^N \times 2^N$ voxels, denoted as $\mathcal{S} = \{0, 1, ..., 2^N-1\}^3$. A point cloud $\mathcal{P}$ is a collection of 3D voxel in this voxelized space $\mathcal{S}$, where each point is defined by its spatial coordinates $\bm{x} = (x, y, z) \in \mathcal{S}$. These voxel collectively forms the point cloud geometry $\mathcal{X}$. 
Each voxel in the point cloud $\mathcal{P}$ is associated with a corresponding attribute. In this study, we consider the RGB color as the attribute of each point. For every non-empty voxel, we assign an attribute $\bm{c} \in [0, 1]^3$, which represents the RGB color values normalized to the range $[0, 1]$. We denote the entire point cloud as $\mathcal{P} = \{\mathcal{X}, \mathcal{C}\}$.
PCC aims to compress point cloud data into a bitstream suitable for digital transmission through an encoder, in order to reduce storage space and transmission bandwidth. At the receiver end, the decoder reconstructs the point cloud by decompressing it. The reconstructed point cloud can be represented as $\hat{\mathcal{P}} = \{\hat{\mathcal{X}}, \hat{\mathcal{C}}\}$. 
This process can be defined as finding a mapping $f: \mathcal{P} \to \hat{\mathcal{P}}$. In this study, we use INR $f(\cdot; \Theta)$ to find this potential mapping, where $\Theta$ are the parameters of INR $f$, and $f(\mathcal{P}; \Theta) = \hat{\mathcal{P}}$. Let $\mathcal{E}(\cdot, \cdot)$ denote the error between two point clouds, and let $\mathcal{R}(\cdot, \cdot)$ represent the compression ratio of the reconstructed point cloud relative to the original point cloud. Given a specified compression ratio $R$, this problem can be expressed as 
\begin{align}\label{eq:optim_target}
\underset{\Theta}{\text{argmin}}\ \mathcal{E}(f(\mathcal{P};\Theta), \mathcal{P}), \quad \text{subject to}\ \mathcal{R}(f(\mathcal{P};\Theta), \mathcal{P}) \leq R. 
\end{align}

Our framework leverages an INR to model a continuous mapping from spatially normalized coordinates $(x, y, z) \in [-1, 1]^3$ to both occupancy probability $p \in [0, 1]$ and RGB color attributes $\bm{c} \in [0, 1]^3$. 
However, practical LiDAR-acquired point cloud present a critical challenge: the inherent sparsity of measured 3D points creates a severely imbalanced learning scenario when using a single INR. Under this configuration, the network must simultaneously model both the sparse observed points and the predominantly empty space, leading to redundant predictions for most query locations while introducing significant optimization challenges.

To address this challenge, we introduce a two-stage decomposition of the mapping $f$ in Equation~\ref{eq:optim_target} into geometry-aware ($f_{g}$) and attribute-aware ($f_{a}$) components, formulated as:
\begin{align}
&f_{g}: \mathcal{X} \to \hat{\mathcal{X}}, \label{eq:fg_o}\\
&f_{a}: \texttt{AttributeMapping}(\hat{\mathcal{X}}, \mathcal{X}, \mathcal{C}) \to \hat{\mathcal{C}}\label{eq:fa_o}.
\end{align}
Two specialized INRs are independently optimized through dedicated objective functions to approximate respective sub-functions, where the \texttt{AttributeMapping} module establishes correspondences between $\mathcal{C}$ and $\hat{\mathcal{X}}$ through nearest-neighbor association.

This separation transforms the original problem into two well-defined sub-tasks: \textbf{(a)} imbalanced classification in a dense voxelized space, and \textbf{(b)} value regression conditioned on sparse surface samples. The decomposition explicitly disentangles the conflicting objectives of modeling spatial density distributions and continuous attribute fields, thereby substantially reducing optimization complexity. Implementation specifics for both geometry and attribute compression will be detailed in subsequent sections.

\subsection{Geometry Compression}

\subsubsection{Problem Modeling}

We define geometry compression as an imbalanced classification problem in a dense voxelized space. Since we employ INR to address this problem, directly implementing $f_{g}: \mathcal{X} \to \hat{\mathcal{X}}$ is not feasible. Therefore, we modify Equation~\ref{eq:fg_o} to 
\begin{align}
&f_{g}: (x, y, z) \in [-1, 1]^3 \to p \in [0, 1], \label{eq:fg} \\ 
&\hat{\mathcal{X}} = \{\bm{x} \mid f_g(\bm{x}') > \tau_0, \bm{x}' = \bm{x}/2^{N-1}-1, \bm{x} \in \mathcal{V}\}. \label{eq:fg_x_bar}
\end{align}
We normalize the voxel coordinates within the optimized sampling voxelized space $\mathcal{V}$ as the input to the $f_{g}$, which outputs the occupancy probability $p$. We dynamically optimize the threshold $\tau_0$ and consider all voxels with $p > \tau_0$ as non-empty, with these voxels contributing to the reconstructed geometry $\hat{\mathcal{X}}$. The methodology for determining the optimized threshold $\tau_0$ and the sampling space $\mathcal{V}$ will be elaborated in subsequent sections.

\subsubsection{Sampling Strategy}\label{sec:sampling_strategy}

As described in Section~\ref{sec:preliminaries}, the entire volumetric space $\mathcal{S}$ contains $2^N \times 2^N \times 2^N$ voxels. However, due to the sparse nature of point clouds, the vast majority of voxels are empty. This makes training the model on all voxels both time-consuming and challenging. 
We redefine the voxelized space with an $M$-bit resolution (where $M < N$), partitioning it into $2^M \times 2^M \times 2^M$ cubes. Each cube $w$ contains $2^{N-M} \times 2^{N-M} \times 2^{N-M}$ voxels. During both training and inference phases, our method exclusively processes the set of non-empty cubic regions $\mathcal{W}$. The optimized sampling space $\mathcal{V}$ is defined as the union of all fine-grained voxel coordinates governed by these cubes:
\begin{align}
    \mathcal{W} = \{\bm{w} \mid \bm{w} = \lfloor \bm{x}/2^{N-M} \rfloor, \bm{x} \in \mathcal{X}\}, \\
    \mathcal{V} = \{\bm{x} \mid \lfloor \bm{x}/2^{N-M} \rfloor \in \mathcal{W}, \bm{x} \in \mathcal{S}\}.
\end{align}

While reducing the sampling space from $\mathcal{S}$ to $\mathcal{V}$ eliminates substantial empty voxels, non-empty voxels still constitute only a minimal proportion $\delta$ within $\mathcal{V}$. This severe class imbalance poses significant training challenges. To address this, we propose a reweighted sampling strategy that maintains a target non-empty voxel ratio of $\alpha=0.5$ within each training batch $\mathbf{x}$. 
Let $\mathcal{U}(\cdot)$ denote uniform sampling over a space. Our sampling strategy would be
\begin{align}
\mathbf{x} &= \alpha\, \mathcal{U}(\mathcal{X}) + (1-\alpha)\, \mathcal{U}(\mathcal{V} - \mathcal{X}),
\end{align}
However, explicitly storing and processing empty voxels $\mathcal{V} - \mathcal{X}$ incurs prohibitive computational and memory overhead. 
To address this, we sample from $\mathcal{W}$ with localized bias $w'$ terms per cube $w$, effectively approximating sampling across $\mathcal{V}$. Our strategy becomes 
\begin{align}
\mathbf{x} &= \hat{\alpha}\, \mathcal{U}(\mathcal{X}) + (1-\hat{\alpha})\, \mathcal{U}(\mathcal{V}),
\end{align}
where the calibrated sampling rate $\hat{\alpha}$ is defined through $\hat{\alpha} = (\alpha - \delta)/(1 - \delta)$, preserving the target class ratio.

\subsubsection{Dynamic Threshold}\label{sec:dynamic_threshold}

To address threshold selection for distinguishing empty and non-empty voxels, we propose an adaptive thresholding strategy. Given the spatial adjacency of empty and non-empty voxels in point cloud data, we formulate the network outputs as $\mathcal{O} = \{p \mid f_g(\bm{x}'), \bm{x}' = \bm{x}/2^{N-1} - 1, \bm{x} \in \mathcal{V}\}$ and introduce a quality metric $\mathcal{D}(\mathcal{O}, \tau)$, measured by D1 PSNR. Empirical verification confirms that $\mathcal{D}$ generally exhibits unimodality with respect to $\tau$, allowing efficient optimization of $\tau_0$ through golden section search to maximize geometric fidelity in Equation~\ref{eq:fg_x_bar}.

\subsection{Attribute Compression}

We formulate attribute compression as continuous regression over sparse coordinate samples. In our approach, each coordinate sample is mapped to its corresponding attribute value through an INR $f_a$. To implement this idea, we redefine the operator introduced in Equation~\ref{eq:fa_o} as follows:
\begin{align}
&f_{a}: (x, y, z) \in [-1, 1]^3 \to \bm{c} \in [0, 1]^3, \label{eq:fa} \\ 
&\hat{\mathcal{C}} = \{f_a(\bm{x}') \mid \bm{x}' = \bm{x}/2^{N-1}-1, \bm{x} \in \hat{\mathcal{X}}\}. \label{eq:fa_x_bar}
\end{align}
To align these predictions with the original point cloud, we project the original attributes $\mathcal{C}$ onto the reconstructed geometry $\hat{\mathcal{X}}$ using a nearest-neighbor mapping strategy. Specifically, for each point $\hat{\bm{x}}_i$ in the reconstructed geometry, we identify the closest point $\bm{x}_k$ in the original geometry $\mathcal{X}$ and assign its attribute value $\bm{c}_j$ to $\hat{\bm{x}}_i$:
\begin{align}
\tilde{\mathcal{C}} = \{\, \bm{c}_j \mid j = \mathop{\text{argmin}}\limits_{k} \|\hat{\bm{x}}_i - \bm{x}_k\|_2, \ \hat{\bm{x}}_i \in \hat{\mathcal{X}}, \ \bm{x}_k \in \mathcal{X}\}.
\end{align}
This nearest-neighbor mapping produces $\tilde{\mathcal{C}}$, which serves as ground truth for the attribute regression task. This optimization process ensures that the attribute INR $f_a$ accurately captures the underlying attribute distributions present in the original point cloud.

\subsection{Learnable Activation Function Network}

\begin{figure}[t]
  \centering
  \includegraphics[width=\linewidth]{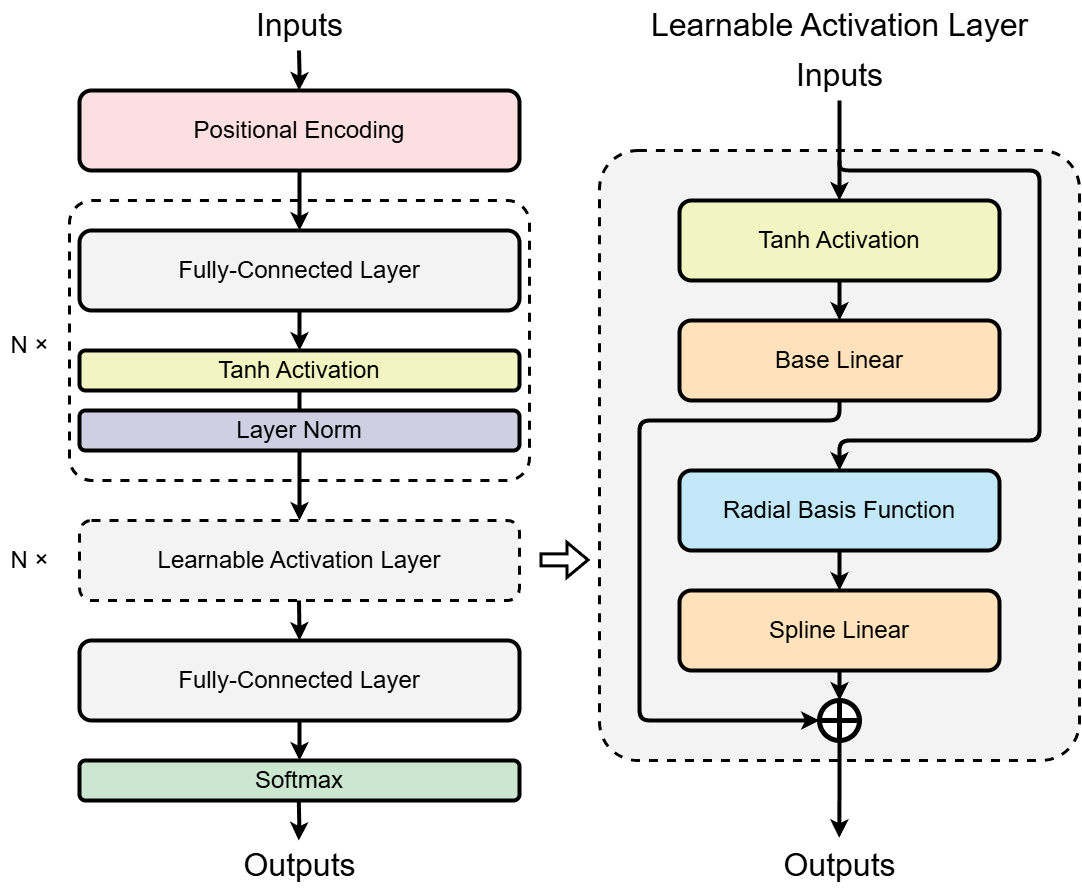}
  \caption{
  Model Architecture of LeAFNet. 
  %
  }
  \Description{LeAFNet utilizes fully connected layers to transform the input, augmented with positional encoding, into a latent space. It employs a Learnable Activation Layer to learn representations, and finally produces the output through fully connected layers.}
  \label{fig:model}
\end{figure}

We propose an INR backbone tailored for PCC, \textbf{LeAFNet} (\textsc{\underline{Le}arnable} \textsc{\underline{A}ctivation} \textsc{\underline{F}unction} \textsc{\underline{Net}work}). Compared to conventional MLPs, it is better at fitting implicit functions in both geometric and attribute compression. We provide an illustration of the network structure in Figure~\ref{fig:model}, and a detailed description will be presented in the subsequent sections.

\subsubsection{Positional Encoding}
Our architecture processes 3D voxel coordinates through positional embedding to enhance spatial correlation modeling. Following NeRF-style positional encoding~\cite{mildenhall2021nerf}, we project inputs into $(6L+3)$-dimensional feature space via Equation~\ref{eq:positional_encode}, where $L$ controls spectral resolution. This frequency-based expansion improves the network's capacity to represent high frequency geometry and attribute patterns critical for precise voxel reconstruction.
\begin{align}\label{eq:positional_encode}
    \Gamma(\bm{x}; L) = (x, &\sin(2^0\pi\bm{x}), \cos(2^0\pi\bm{x}),\ ...\ , \notag\\
    &\sin(2^{L-1}\pi\bm{x}), \cos(2^{L-1}\pi\bm{x}) )
\end{align}

Positional encoding employs sinusoidal basis functions across octave frequencies along each coordinate axis. This spectral decomposition enables coordinate-conditioned networks to resolve high-frequency geometric signals critical for reconstruction fidelity in compression tasks. The resulting multi-scale descriptors encode non-local geometric relationships through expanded feature dimensions, effectively addressing the limited representational capacity of raw coordinate inputs.

\subsubsection{Learnable Activation Layer}

LeAFNet's architectural objective centers on learnable activation operators that enhance implicit function approximation while maintaining parameter efficiency. Targeting point cloud compression, we adopt a hierarchical structure: positional encoding maps inputs to latent features through fully-connected layers, followed by cascaded activation modules with parametric nonlinearities. Crucially, our activation design replaces traditional B-spline kernels with radial basis functions (RBFs) inspired by recent spectral networks~\cite{li2024kolmogorov}. This can be expressed as
\begin{align}\label{eq:kan_rbf}
    &\phi(x) = w_b\texttt{silu}(x) + w_s\sum^N_{i=1}\text{exp}(-\frac{\lVert x-c_i \rVert^2}{h^2}).
\end{align}
This design significantly accelerates both the forward and backward passes, enabling faster computation and stable training.

While KAN-inspired architectures underperform MLPs in standard vision tasks, their adaptive activation mechanisms prove particularly effective for INR-based compression. Our empirical observations reveal that learnable activation layers achieve comparable accuracy to MLPs with shallower networks (reducing $N$ and $L$ by 4-8 times), offsetting their per-layer parameter overhead. This parameter-accuracy trade-off aligns with PCC's requirement for compact yet expressive models: MLPs require $O(N^2L)$ parameters versus $O(N^2LG)$ for activation layers~\cite{liu2025kan}, but practical implementations favor our approach due to drastically reduced depth and width requirements. These findings position LeAFNet as a parameter-efficient solution for PCC.

\subsection{PICO}

In this section, we will introduce \textbf{PICO}, an INR-based PCC algorithm, and how the previously introduced geometry and attribute compression are applied as modules throughout the entire pipeline. The overall PICO algorithm is presented in Algorithm~\ref{alg:pico_pipeline}.

\subsubsection{Loss Function}

\begin{algorithm}[t]
\caption{PICO: INR-based Point Cloud Compression}
\label{alg:pico_pipeline}
\begin{algorithmic}[1]  
    \Require Point Cloud $\mathcal{P} = \{\mathcal{X}, \mathcal{C}\}$, Model Dictionary $\mathcal{M}$
    
    \Statex
    \Statex \textbf{Compression Stage}
    \State $\mathcal{W} \gets \{\bm{w} \mid \bm{w} = \lfloor \bm{x}/2^{N-M} \rfloor, \bm{x} \in \mathcal{X}\}$
    \State $\mathcal{V} \gets \{\bm{x} \mid \lfloor \bm{x}/2^{N-M} \rfloor \in \mathcal{W}, \bm{x} \in \mathcal{S}\}$ 

    \State $\Theta_g^{(0)}, \Theta_a^{(0)} \gets \texttt{SelectModel}(\mathcal{M}, b)$ 
    
    \Statex
    \For{$t = 0$ to $T_g$} \Comment{Training geometry INR}
        \State $\textbf{x} \gets \texttt{Sample}(\mathcal{V})$
        \State $\Theta_g^{(t+1)} \gets \Theta_g^{(t)} - \gamma \nabla \mathcal{L}_\text{geometry}(f_g(\textbf{x}; \Theta_g^{(t)}))$
    \EndFor
    
    \State $\hat{\Theta}_g \gets \texttt{Quantization}(\Theta_g^{(T_g)}, \delta_g)$
    
    \State $\mathcal{O} \gets \{p \mid f_g(\bm{x}'; \hat{\Theta}_g), \bm{x}' = \bm{x}/2^{N-1} - 1, \bm{x} \in \mathcal{V}\}$
    
    \State $\tau_0 \gets \texttt{AdaptiveThreshold}(\mathcal{O})$
    
    \State $\hat{\mathcal{X}} \gets \{\bm{x} \mid f_g(\bm{x}'; \hat{\Theta}_g) > \tau, \bm{x}' = \bm{x}/2^{N-1}-1, \bm{x} \in \mathcal{V}\}$
    
    \Statex
    \For{$t = 0$ to $T_a$} \Comment{Training attribute INR}
        \State $\textbf{x} \gets \texttt{Sample}(\hat{\mathcal{X}})$
        \State $\Theta_a^{(t+1)} \gets \Theta_a^{(t)} - \gamma \nabla \mathcal{L}_\text{attribute}(f_a(\textbf{x}; \Theta_a^{(t)}))$
    \EndFor

    \State $\hat{\Theta}_a \gets \texttt{Quantization}(\Theta_a^{(T_a)}, \delta_a)$
    
    \State $\widetilde{\Theta}_g, \widetilde{\Theta}_a \gets \texttt{EntropyEncode}(\hat{\Theta}_g, \hat{\Theta}_a)$ 
    
    \Statex
    \Statex \textbf{Decompression Stage}
    \State $\hat{\Theta}_g, \hat{\Theta}_a \gets \texttt{EntropyDecode}(\widetilde{\Theta}_g, \widetilde{\Theta}_a)$ 
    
    \State $\hat{\mathcal{X}} \gets \{\bm{x} \mid f_g(\bm{x}'; \hat{\Theta}_g) > \tau, \bm{x}' = \bm{x}/2^{N-1}-1, \bm{x} \in \mathcal{V}\}$
    
    \State $\hat{\mathcal{C}} \gets \{f_a(\bm{x}'; \hat{\Theta}_a) \mid \bm{x}' = \bm{x}/2^{N-1}-1, \bm{x} \in \hat{\mathcal{X}}\}$
    
    \State $\hat{\mathcal{P}} \gets \{\hat{\mathcal{X}}, \hat{\mathcal{C}}\}$

    \Statex \Return $\hat{\mathcal{P}}$
\end{algorithmic}
\end{algorithm}

Following our problem formulation in Section~\ref{sec:preliminaries}, we formalize PCC as dual learning objectives with task-specific loss formulations to guide network optimization. 
For geometric compression, we implement an $\alpha$-modulated focal loss. Building upon the standard focal loss~\cite{lin2017focal}, which addresses class imbalance through its adaptive weighting mechanism that prioritizes hard examples via a $(1-p_t)^\gamma$ modulator, we introduce additional $\alpha$-balancing in Equation~\ref{eq:loss_function_g} to handle severe label imbalance.
Attribute reconstruction employs voxel-wise MSE loss computed exclusively over occupied voxels, enforcing perceptual color accuracy while maintaining geometric consistency.

Unlike prevailing codec architectures that regulate bitrate through entropy model priors, PICO reformulates PCC as a neural network compression problem. We control the sparsity of model parameters via $\ell_1$ regularization, thereby governing the bitstream size obtained through model quantization and entropy coding. This is implemented by incorporating an $\ell_1$ regularization term into our loss function. The two loss functions in geometry and attribute compression we design are formulated as follows:
\begin{align}
\mathcal{L}_{\text{geometry}} &= -\mathbb{E}_{i \sim \mathcal{X}} \big[\alpha (1-p_i)^\gamma \log p_i \cdot \mathbb{I}(y_i=1) \notag \\
 &+ (1-\alpha) p_i^\gamma \log(1-p_i) \cdot \mathbb{I}(y_i=0) \big] + \lambda_g \|\Theta_g\|_1 \label{eq:loss_function_g}\\
\mathcal{L}_{\text{attribute}} &= \mathbb{E}_{i \sim \tilde{\mathcal{C}}} \big[ \|\bm{c}_i - \hat{\bm{c}}_i\|_2^2 \big] + \lambda_a \|\Theta_a\|_1 \label{eq:loss_function_a}
\end{align}

\subsubsection{Adaptive Model Parameter Selection}\label{sec:model_selection}

Models with reduced parameter counts exhibit clear limitations in matching the compression performance of larger counterparts under high bits-per-point (bpp) regimes. Notably, as bpp decreases, these compact architectures demonstrate delayed onset of the abrupt distortion degradation observed in larger models. This phenomenon motivates a practical guideline: lightweight models should be prioritized for high compression ratios, while capacity-rich architectures better serve scenarios demanding superior point cloud fidelity.

The envelope formed by rate-distortion (RD) curves across all models spans a continuous parameter space, theoretically forming the optimal RD curve for compression tasks. While an idealized system could dynamically select models with varying parameter counts to optimize performance at any target bpp $b$, such continuous adaptation remains impractical. Instead, we propose a model dictionary $\mathcal{M}$ that enables adaptive selection of optimal architectures based on target compression rates or bpp thresholds, establishing an efficient and flexible framework for real-world deployment.

\begin{table*}[t]
\centering
\caption{
Static PCC Performance of PICO (MLP), G-PCC (octree), G-PCC (trisoup) and V-PCC against PICO. 
We conduct static PCC experiments on the first frame of four point cloud sequences from \texttt{8iVFB}. Using the Bjontegaard delta metric to calculate the gains of PICO over baseline methods, we observe that PICO achieves highly competitive results in both geometry compression and joint geometry and attribute compression.
}
\label{tab:static_baseline_geometry}
\begin{tabular}{ccccccccc}
\toprule
\textbf{Point Cloud} & \multicolumn{2}{c}{\textbf{PICO (MLP) (Ours)}} & \multicolumn{2}{c}{\textbf{G-PCC (octree)}} & \multicolumn{2}{c}{\textbf{G-PCC (trisoup)}} & \multicolumn{2}{c}{\textbf{V-PCC}}\\
\midrule
Geometry Only & \makecell[c]{BD-BR} & \makecell[c]{BD-PSNR} & \makecell[c]{BD-BR} & \makecell[c]{BD-PSNR} & \makecell[c]{BD-BR} & \makecell[c]{BD-PSNR} & \makecell[c]{BD-BR} & \makecell[c]{BD-PSNR} \\
\cmidrule(lr){1-9}
\texttt{longdress}      & -33.13 & 2.80 & -69.19 & 7.33 & -49.34 & 3.96 & -35.42 & 2.64 \\
\texttt{loot}           & -45.53 & 4.56 & -71.78 & 7.84 & -53.33 & 4.36 & -44.66 & 3.64 \\
\texttt{redandblack}    & -48.12 & 4.92 & -69.25 & 6.48 & -47.20 & 3.33 & -53.27 & 4.42 \\
\texttt{soldier}        & -43.55 & 4.36 & -60.76 & 6.69 & -42.75 & 4.15 & -45.51 & 4.20 \\
\textbf{\texttt{average}} & \textbf{-42.58} & \textbf{4.16} & \textbf{-67.75} & \textbf{7.09} & \textbf{-48.16} & \textbf{3.95} & \textbf{-44.72} & \textbf{3.73} \\
\midrule
Attribute \& Geometry & \makecell[c]{BD-BR} & \makecell[c]{BD-PCQM} & \makecell[c]{BD-BR} & \makecell[c]{BD-PCQM} & \makecell[c]{BD-BR} & \makecell[c]{BD-PCQM} & \makecell[c]{BD-BR} & \makecell[c]{BD-PCQM} \\
\cmidrule(lr){1-9}
\texttt{longdress}      & -17.19 & 1.35 & -40.94 & 3.89 & -19.88 & 0.88 &   8.19 & -0.68 \\
\texttt{loot}           & -49.41 & 2.51 & -74.72 & 6.86 & -47.68 & 1.43 & -29.53 & 0.22 \\
\texttt{redandblack}    & -51.77 & 2.51 & -71.68 & 5.86 & -40.30 & 1.73 & -46.48 & 1.15 \\
\texttt{soldier}        & -43.68 & 3.06 & -65.87 & 7.69 & -57.49 & 2.37 & -26.18 & 1.08 \\
\textbf{\texttt{average}} & \textbf{-40.51} & \textbf{2.36} & \textbf{-63.30} & \textbf{6.08} & \textbf{-41.34} & \textbf{1.60} & \textbf{-23.50} & \textbf{0.44}\\
\bottomrule
\end{tabular}
\end{table*}
\begin{figure*}[t]
  \centering
  \includegraphics[width=\linewidth]{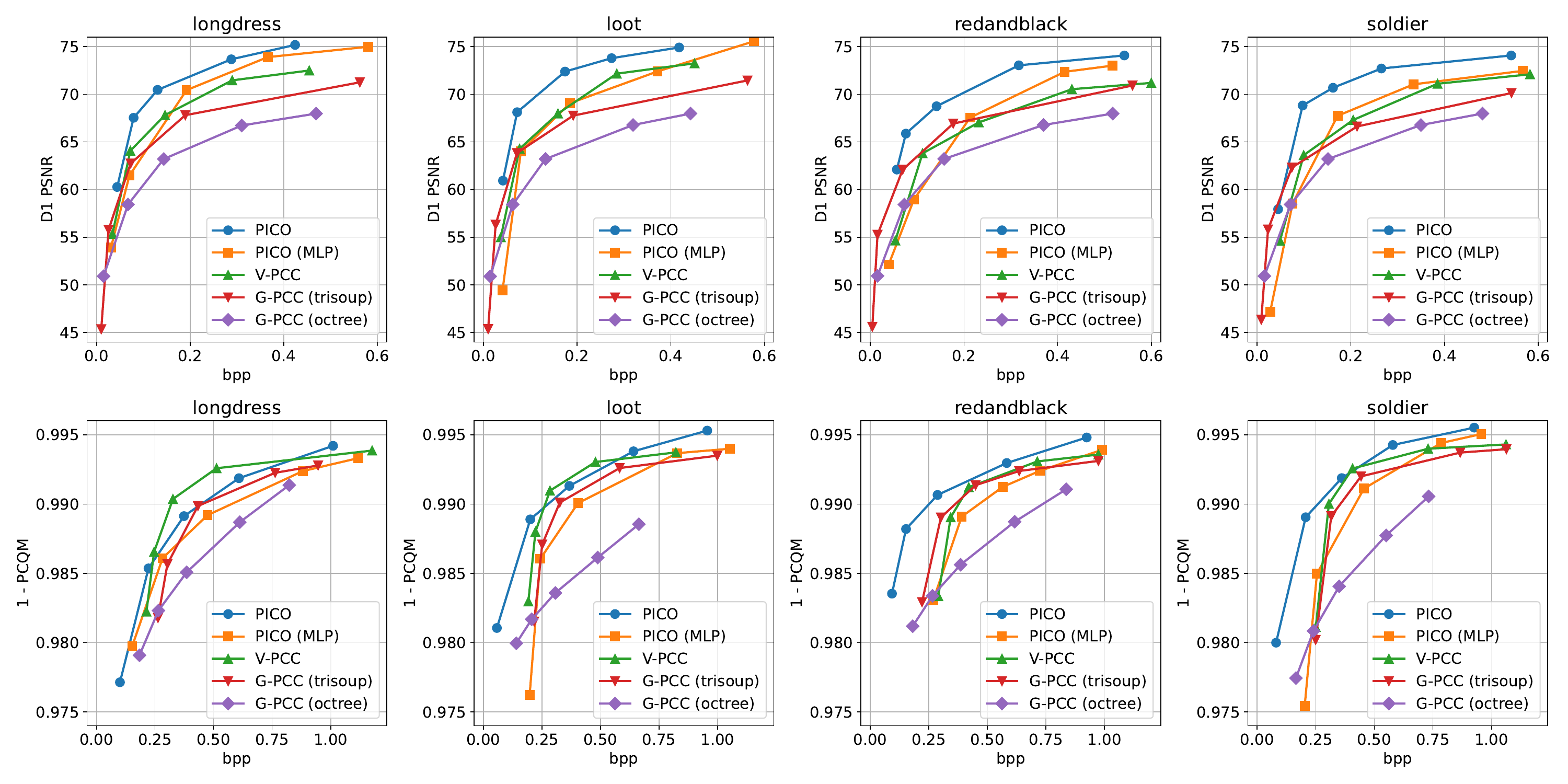}
  \caption{Rate-distortion curves of PICO, PICO (MLP), G-PCC (octree), G-PCC (trisoup) and V-PCC on Static PCC.
  %
  } 
  \Description{
  }
  \label{fig:baseline_experiments}
\end{figure*}

\subsubsection{Quantization \& Entropy Coding}

By encoding point cloud data through INR, we systematically reformulate PCC as a neural network compression task. While $\ell_1$-regularized training establishes the foundation, the subsequent compression pipeline involves quantizing model parameters using step sizes $\Delta_g$ and $\Delta_a$, followed by entropy coding to produce efficiently compressed model bitstreams. Our framework specifically employs DeepCABAC~\cite{wiedemann2019deepcabac}, a state-of-the-art context-adaptive binary arithmetic coder specialized for deep neural network compression. Through joint optimization of the regularization parameter $\lambda$ and quantization step $\Delta$, PICO achieves precise rate control by systematically balancing bitstream size against reconstruction fidelity.
\section{Experiments}

\subsection{Settings}

We introduce the datasets and baselines used in our study. Additionally, we provide details on the experimental setup and the evaluation metrics. 

\subsubsection{Dataset}

We conduct our experiments on 8i Voxelized Full Bodies (8iVFB). The dataset contains four sequences: \texttt{longdress}, \texttt{loot}, \texttt{redandblack}, and \texttt{soldier}. Each sequence captures the full body of a human subject using 42 RGB cameras arranged in 14 clusters, with each cluster acting as a logical RGBD camera. The capture rate is 30 fps over a 10-second period. For each sequence, a single spatial resolution is provided: a 1024x1024x1024 voxel cube, referred to as depth 10. In each cube, only the voxels near the surface of the subject are occupied. The attributes of an occupied voxel correspond to the red, green, and blue components of the surface color. 
The cube is scaled for each sequence to form the smallest bounding cube that contains the entire sequence. Since the subject's height is typically the longest dimension, for a subject with a height of $1.8$ m, each voxel at depth 10 corresponds to approximately $1.8\ \text{m} / 1024\ \text{voxels} \approx 1.75\ \text{mm}$ per voxel on each side.

\subsubsection{Baselines}
We adopt two methods from the MPEG standard as our baselines: Geometry-based Point Cloud Compression (G-PCC)\fthref{https://github.com/MPEGGroup/mpeg-pcc-tmc13}~\cite{graziosi2020overview} and Video-based Point Cloud Compression (V-PCC)\fthref{https://github.com/MPEGGroup/mpeg-pcc-tmc2}~\cite{graziosi2020overview}. G-PCC is evaluated using its two primary variants: G-PCC (octree) and G-PCC (trisoup), each employing different spatial partitioning techniques to encode geometry efficiently. Additionally, for attribute encoding, G-PCC utilizes the Region-Adaptive Hierarchical Transform (RAHT)~\cite{de2016compression}, a widely adopted approach for compressing voxel-based attributes.

\subsubsection{Experimental Detail}
We set $M = 5$ to redefine the voxelized space, where each 3D space cube contains $32\times32\times32$ voxels. We use the Adam optimizer~\cite{kingma2014adam} to train the network, with an initial learning rate of $ \text{lr} = 1 \times 10^{-3}$, which decays by a factor of 0.1. The batch size is set to 32,768. For geometric compression, we set the sampling coefficient $\alpha = 0.5$, the focal loss modulation coefficient $\gamma = 2$. 
In the geometry compression, we perform 120,000 training steps. The model parameters are quantized with a step size $\Delta_g = 1/2^{10}$ after training. 
In attribute compression, we perform 90,000 training steps, and the trained model parameters are quantized with a step size of $\Delta_a = 1/2^{12}$. 
Finally, we use DeepCABAC~\cite{wiedemann2019deepcabac} for lossless compression of the quantized parameters.

\subsubsection{Metrics}

For geometric compression, we measure geometric distortion using the peak signal-to-noise ratio (PSNR) based on point-to-point error (D1), following the common test conditions of MPEG PCC. Formally, the point-to-point error of a point cloud $\mathcal{B}$ relative to a reference point cloud $\mathcal{A}$ is defined as the error calculated in two directions, yielding $e(\mathcal{B}, \mathcal{A})$ and $e(\mathcal{B}, \mathcal{A})$. The D1 PSNR between $\mathcal{B}$ and $\mathcal{A}$ is then given by 
\begin{align}
    e(\mathcal{B}, \mathcal{A}) &= \frac{1}{|\mathcal{B}|}\sum_{\bm{b}\in\mathcal{B}}\min_{\bm{a}\in\mathcal{A}}||\bm{b}-\bm{a}||^2_2, \\
    \text{D1 PSNR} &= 10 \log_{10} \frac{3 \times (2^N - 1)^2}{\max \{e(\mathcal{B}, \mathcal{A}), e(\mathcal{A}, \mathcal{B})\}} (\text{dB}).
\end{align}

For joint geometry and attribute compression, we use PCQM\fthref{https://github.com/MEPP-team/PCQM}~\cite{meynet2020pcqm}, a full-reference quality metric for colored 3D point clouds, as our evaluation metric. This metric quantifies the overall distortion of both geometry and attributes, providing a measure that aligns more closely with the human visual system. Given that a lower PCQM value indicates higher perceptual quality of the point cloud, we use $1-\text{PCQM}$ as the practical evaluation metric.
To compare performance, we utilize the Bjontegaard delta metrics\footnote{\href{https://github.com/tbr/bjontegaard_etro}{https://github.com/tbr/bjontegaard\_etro}}~\cite{bjontegaard2001calculation} to quantify the RD performance gains of different methods.

\subsection{Static Point Cloud Compression}
We evaluated PICO against G-PCC and V-PCC on the 8iVFB dataset using the first frame of four static sequences. RD curves in Figure~\ref{fig:baseline_experiments} and objective metrics (BD-BR, BD-PSNR, BD-PCQM) in Table~\ref{tab:static_baseline_geometry} demonstrate PICO’s compression gains.

In terms of geometric compression performance, PICO achieves an average BD-BR reduction of 67.75\% compared to G-PCC (octree), 48.16\% compared to G-PCC (trisoup), and 44.72\% compared to V-PCC. Furthermore, PICO improves BD-PSNR by 7.09 dB, 3.95 dB, and 3.73 dB, respectively, demonstrating its superior geometric compression efficiency. 
For geometry and attribute joint compression, PICO also exhibits significant performance improvements. Compared to G-PCC (octree), PICO reduces BD-BR by 63.30\%, while achieving reductions of 41.34\% and 23.50\% relative to G-PCC (trisoup) and V-PCC, respectively. Additionally, the BD-PCQM results further validate the advantages of PICO, with improvements of $6.08\times10^{-3}$, $1.60\times10^{-3}$, and $0.44\times10^{-3}$, indicating that PICO better preserves the perceptual quality of point clouds. 

\begin{figure*}[t]
  \centering
  \includegraphics[width=\linewidth]{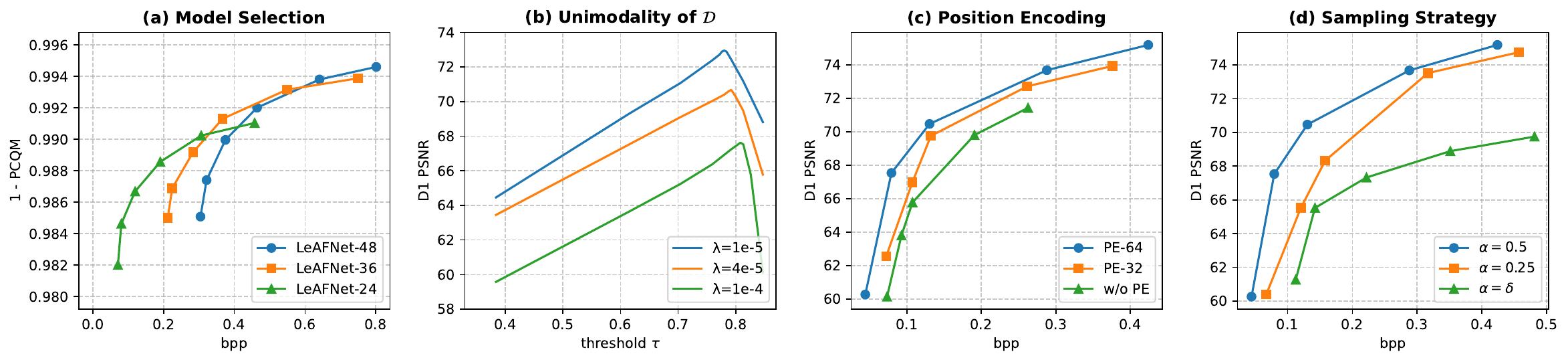}
  \caption{
  (a) PCQM RD curve of LeAFNet with different parameter sizes. (b) Unimodal property of $\mathcal{D}$. (c) PSNR RD curve of LeAFNet with different positional encoding lengths $L$. (d) Impact of different sampling strategies on compression performance.
  %
  }
  \label{fig:ablation_study}
\end{figure*}

\subsection{Ablation Study}

\subsubsection{Impact of LeAFNet}

In LeAFNet, we replace conventional MLP components in INR with learnable activation layers. To investigate the performance advantages of this design, we conduct an ablation study where the learnable activation layers are substituted by residual connection modules consisting of two fully connected layers. Notably, the MLP-base model in this ablation study requires 50\% more parameters than our baseline LeAFNet configuration to achieve comparable performance, demonstrating the parameter efficiency of our proposed architecture. Detailed comparisons are presented in Table~\ref{tab:static_baseline_geometry} and Figure~\ref{fig:baseline_experiments}.

Experimental results demonstrate that LeAFNet delivers superior INR performance with improved parameter efficiency. For geometry compression, the proposed method achieves a 42.58\% BD-BR reduction and a 4.16 dB improvement in BD-PSNR. In the joint compression, we observe 40.51\% BD-BR reduction along with a $2.36\times10^{-3}$ improvement in BD-PCQM. 

\subsubsection{Adaptive Model Parameter Selection}

As discussed in Section~\ref{sec:model_selection}, the theoretical optimal RD curve of PICO could be derived from the convex hull of continuous model parameter selections under ideal conditions. In practical implementations, we instead regulate model capacity by adjusting the hidden dimension of learnable activation layers. To validate this, we conduct ablation studies on the \texttt{Longdress} sequence using three configurations with hidden dimensions 48, 36, and 24 for static PCC. The corresponding PCQM RD curves in Figure~\ref{fig:ablation_study} (a) reveal two critical observations: \textbf{(a)} Smaller models gradually outperform larger counterparts as target bpp decreases, and \textbf{(b)} They exhibit delayed performance degradation caused by $\ell_1$ regularization.

These findings motivate our key design principle. Rather than solely tuning $\ell_1$ regularization intensity, optimal compression should adaptively select model structures from a predefined dictionary $\mathcal{M}$ based on target bpp. This strategy not only enhances compression efficiency but also eliminates redundant computational overhead through training a model of appropriate size.

\subsubsection{Unimodality of $\mathcal{D}$ and Dynamic Threshold}

In Section~\ref{sec:dynamic_threshold}, we introduce a dynamic threshold selection strategy to enhance reconstructed point cloud geometry. The proposed golden section search algorithm identifies optimal $\tau_0$ under the core assumption that $\mathcal{D}(\mathcal{O}, \tau)$ exhibits unimodal behavior. We validate this assumption through empirical analysis across three models trained with varying $\ell_1$ regularization strengths. As demonstrated in Figure~\ref{fig:ablation_study} (b), the D1 PSNR curve under evolving $\tau$ values confirm the prevalence of unimodal characteristics in $\mathcal{D}(\mathcal{O}, \tau)$, enabling effective application of golden-section search for quality enhancement. Quantitative analysis shows our dynamic threshold strategy achieves over $2+$ dB D1 PSNR improvement compared to static threshold baselines, validating its efficacy in geometric refinement.

\subsubsection{Positional Encoding}

The positional encoding module serves as a critical component in LeAFNet, enhancing low dimensional inputs to improve spatial correlation understanding. This encoding strategy enables superior capture of geometric structures and spatial dependencies within point clouds through explicit coordinate embedding. Our default configuration employs $L=64$ for effective positional feature representation. To systematically evaluate its impact, we conduct ablation studies comparing three variants: the baseline ($L=64$), a reduced-capacity version ($L=32$), and LeAFNet without positional encoding.

Experimental results in Figure~\ref{fig:ablation_study} (c) demonstrate two key findings: \textbf{(a)} Reducing $L$ constrains the model’s ability to represent fine-grained spatial relationships, leading to measurable reconstruction quality degradation; \textbf{(b)} Complete removal of positional encoding results in substantial performance deterioration (3.38 dB BD-PSNR drop), indicating its fundamental role in maintaining spatial awareness. These observations confirm positional encoding’s importance in preserving structural fidelity.

\subsubsection{Sampling Strategy}


To address the severe imbalance between empty and non-empty voxel distributions discussed in Section~\ref{sec:sampling_strategy}, we employ a systematic resampling strategy that maintains a controlled proportion of non-empty voxels ($\alpha=0.5$). We validate this approach through comprehensive comparisons with two alternative configurations: a reduced sampling ratio ($\alpha=0.25$) and the original unbalanced distribution ($\alpha=\delta$). 

As demonstrated in Figure~\ref{fig:ablation_study} (d), our weighted sampling strategy yields significant quality improvements, achieving a substantial 5.63 dB BD-PSNR gain over the $\alpha=\delta$ baseline implementation. Empirical observations reveal that under the $\alpha=\delta$ configuration, model training becomes prone to catastrophic collapse and fundamentally struggles to learn the underlying occupancy probability distribution. Our sampling strategy effectively mitigates these training instabilities while reducing optimization challenges.
\section{Conclusion}

In this work, we investigate the application of implicit neural representations to point cloud compression. Toward this objective, we develop a two-stage framework, PICO, which decouples point cloud compression into geometry compression and attribute compression. Building on principles from KAN, we design LeAFNet, a network that employs learnable activation functions in the latent space to enhance implicit function approximation. Through rigorous experiments, we verify that our method substantially improves existing compression standards and achieves state-of-the-art performance in benchmark evaluations. We hope that our work will provide new insights for the growing body of deep learning-based encoder-decoder methods, paving the way for novel frameworks and contributing to the development of next-generation deep learning-based point cloud compression standards.

\newpage

\bibliographystyle{ACM-Reference-Format}
\bibliography{reference}
\end{document}